\pdfoutput=1

\documentclass[11pt]{article}

\usepackage[preprint]{acl}

\usepackage{times}
\usepackage{latexsym}

\usepackage[T1]{fontenc}

\usepackage[utf8]{inputenc}

\usepackage{microtype}

\usepackage{inconsolata}

\usepackage{graphicx} 
\usepackage{multirow}
\usepackage{makecell}
\usepackage{enumitem}
\usepackage{amsmath}

\usepackage[ruled,vlined]{algorithm2e}
\usepackage{booktabs}
\usepackage{amsmath}
\usepackage{multirow}
\usepackage{bbding}
\usepackage{mathrsfs}
\usepackage{subfigure}
\usepackage{amssymb}
\usepackage{mathtools}
\usepackage{xcolor}
\usepackage[most]{tcolorbox}
\usepackage{hyperref}

\title{Knowledge Editing on Black-box Large Language Models}

\author{Xiaoshuai Song$^{1*}$, Zhengyang Wang$^{1*}$, Keqing He$^{2}$, Guanting Dong$^{1}$, Yutao Mou$^{1}$\\{\bf Jinxu Zhao$^{1}$, \bf Weiran Xu$^{1}$}\thanks{\ \ The first two authors contribute equally. Weiran Xu is the corresponding author.}\\
  $^1$Beijing University of Posts and Telecommunications, Beijing, China\\
$^{2}$Meituan, Beijing, China\\
  \texttt{\{songxiaoshuai,wzyang,dongguanting,myt,zhaojinxu,xuweiran\}@bupt.edu.cn}\\
  \texttt{hekeqing@meituan.com}
}

\begin{document}
\maketitle
\begin{abstract}
Knowledge editing (KE) aims to efficiently and precisely modify the behavior of large language models (LLMs) to update specific knowledge without negatively influencing other knowledge. Current research primarily focuses on white-box LLMs editing, overlooking an important scenario: black-box LLMs editing, where LLMs are accessed through interfaces and only textual output is available. In this paper, we first officially introduce KE on black-box LLMs and then propose a comprehensive evaluation framework to overcome the limitations of existing evaluations that are not applicable to black-box LLMs editing and lack comprehensiveness. To tackle privacy leaks of editing data and style over-editing in current methods, we introduce a novel postEdit framework, resolving privacy concerns through downstream post-processing and maintaining textual style consistency via fine-grained editing to original responses. Experiments and analysis on two benchmarks demonstrate that postEdit outperforms all baselines and achieves strong generalization, especially with huge improvements on style retention (average $+20.82\%\uparrow$).\footnote{We release our code at \url{ https://github.com/songxiaoshuai/postEdit}.}
\end{abstract}

\section{Introduction}
\begin{figure}[t]
    \centering
    \resizebox{0.48\textwidth}{!}{
    \includegraphics{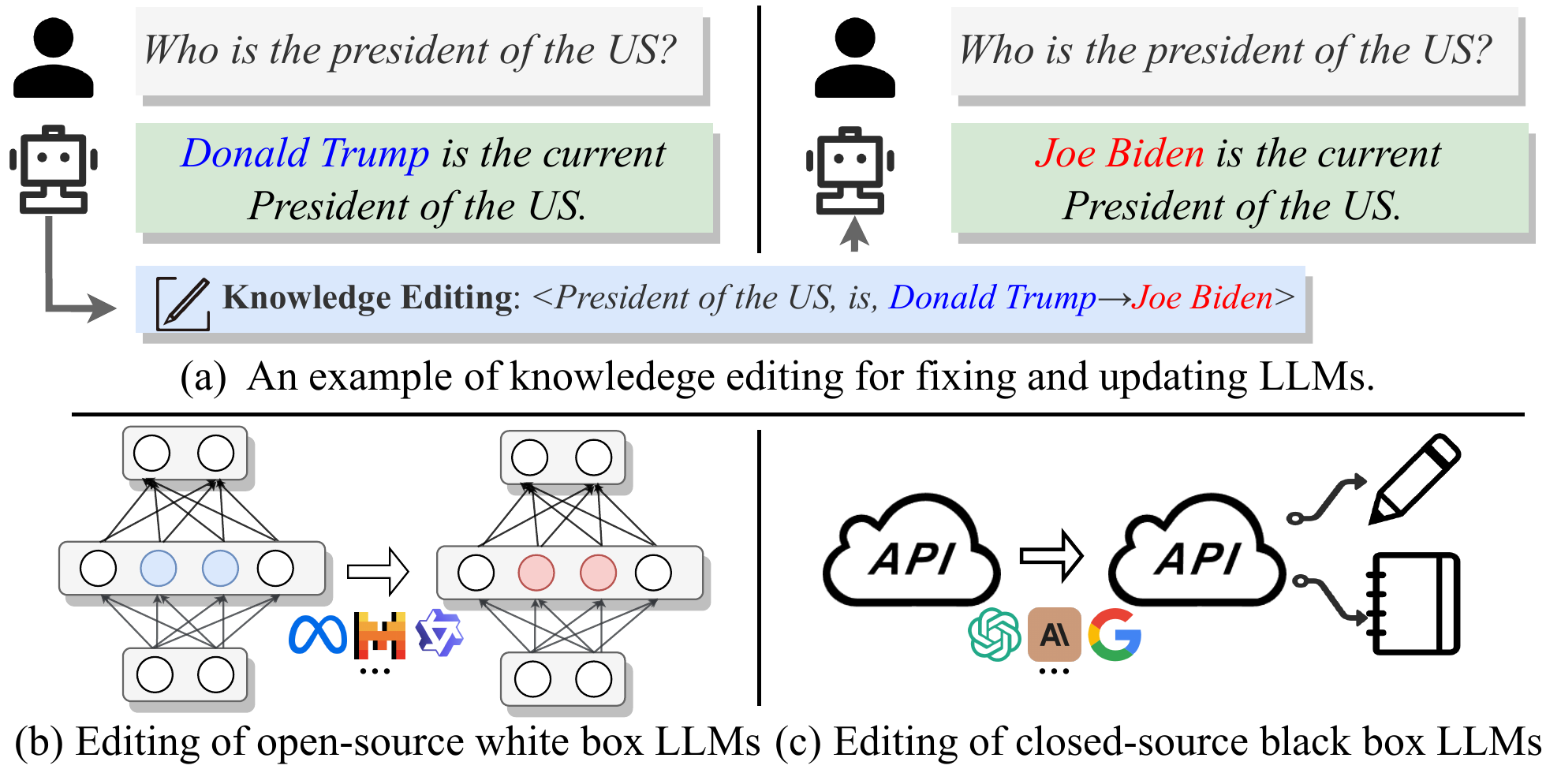}}
    \vspace{-0.8cm}
    \caption{Illustration of Knowledge Editing and comparison of two editing scenarios, where black-box LLMs editing constrains LLMs to only obtain textual output.}
    \vspace{-0.5cm}
    \label{fig:intro_task}
\end{figure}

Recently, large language models (LLMs) have swept through the natural language processing (NLP) community \cite{zhao2023survey,chang2023survey}. Pre-trained on extensive corpora, LLMs acquire substantial real-world knowledge and are utilized for knowledge-intensive tasks \cite{liu2023vera,bian2023chatgpt,wang2023evaluating}. However, as the world's state evolves, the requirement of updating LLMs to rectify obsolete information or incorporate new knowledge to maintain their relevance is constantly emerging. Frequent fine-tuning is impractical due to intensive computational overload and the catastrophic forgetting caused by overfitting to new data \cite{feng2023trends,wang2023knowledge}. To address this issue, the concept of knowledge editing (\textbf{KE}, also known as model editing) has been proposed, aiming to efficiently and precisely modify the behavior of LLMs to update specific knowledge without negatively influencing other knowledge \cite{yao-etal-2023-editing,wang2023knowledge}, as illustrated in Fig \ref{fig:intro_task}(a). 

A prevalent approach to KE involves manipulating the internals of LLMs through gradients or causal analysis \cite{de-cao-etal-2021-editing,mitchell2021fast,meng2022locating,meng2022mass,huang2023transformer}, as depicted in Fig \ref{fig:intro_task}(b).  While these methods have shown promise, they require LLMs to be locally deployed and parameter-transparent, termed white-box LLMs editing. 
In more typical scenarios, LLMs are provided via APIs by upstream manufacturers (e.g., OpenAI, Google) for downstream services, with inaccessible internal workings and text-only output. We refer to KE on such LLMs as \textbf{black-box LLMs editing}, as shown in Fig \ref{fig:intro_task}(c).
This raises a key question: \textit{how can we edit "black-box" models when undesired outputs or errors occur?} Furthermore, existing KE evaluation protocols rely on changes in the model's logits before and after editing, and are unattainable for black-box LLMs, prompting another question: \textit{how can we comprehensively evaluate black-box KE methods?}

\begin{figure*}[t]
    \centering
    \resizebox{0.98\textwidth}{!}{
    \includegraphics{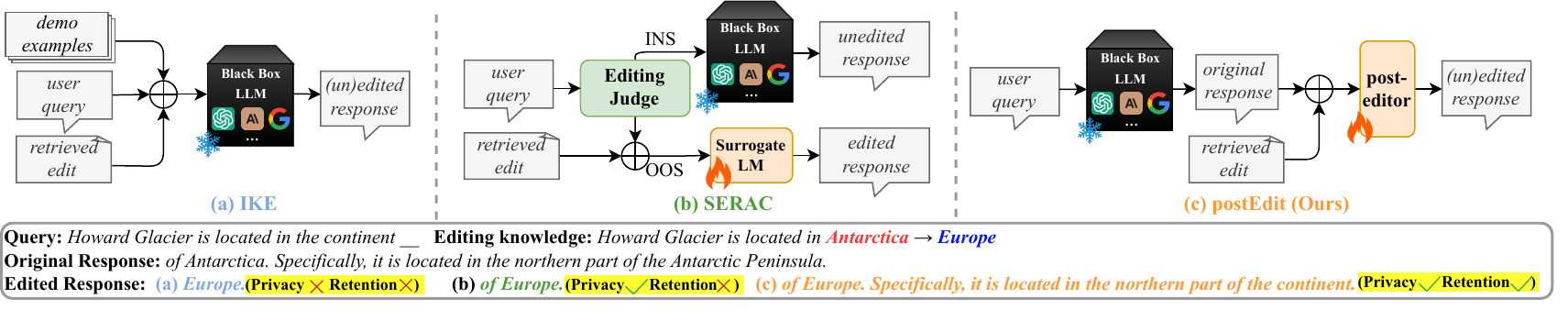}}
    \vspace{-0.35cm}
    \caption{Comparison of different KE frameworks for black-box LLM editing. IKE operates on LLM input, and SERAC performs editing using a surrogate model parallel to LLM, while our postEdit edits after the output of LLM and achieves both privacy protection and style retention.}
    \vspace{-0.35cm}
    \label{fig:intro_contrast}
\end{figure*}

There are some studies based on external memory that can be applied to black-box LLM editing scenarios. SERAC \cite{mitchell2022memory} utilizes an surrogate model to generate edited responses when queries are classified within the editing scope. IKE \cite{zheng-etal-2023-edit} facilitates in-context learning \cite{dong2022survey} of LLM itself by demonstrating exemplars to learn the ability to discern the need of editing and how to edit. However, as depicted in Fig \ref{fig:intro_contrast}(a)(b), these methods encounter two crucial drawbacks: 
(1) \textbf{Privacy leakage of editing data}. IKE inputs recall data from the demonstration library and editing memory to LLMs, inevitably disclosing downstream private editing data to upstream LLM providers. 
(2) \textbf{Style over-editing}.\footnote{In this paper, the style extensively covers the expressive forms, conciseness, length, etc., of the text.} The KE methods should only edit the knowledge of LLMs while keeping the text style unchanged. Specifically, the different scales or types between the surrogate model and base LLM result in stylistic differences for SERAC, while LLM's sensitivity to prompts and demonstrations \cite{chen-etal-2023-relation} leads to style over-editing in IKE. Therefore, even though their edited responses both target the new object "\textit{Europe}", they exhibit a pronounced departure in style from the original responses. An ideal black-box editing method should preserve downstream data privacy while achieving commendable editing performance and style retention.

In this paper, we firstly revisit the existing evaluation of KE and formulate an improved general evaluation framework for black-box LLM editing. In addition to the traditional lexical evaluation of knowledge editing, our framework incorporates the assessment of style retention for the first time and conducts a comprehensive evaluation from both textual and semantic perspectives. (see Section \ref{sec:eval}).
To solve the problems of existing methods mentioned above, we propose a novel post-editing approach termed \textbf{postEdit}, applied after the output of LLMs, as illustrated in Fig \ref{fig:intro_contrast}(c). Diverging from previous approaches, on the one hand, the post-processing mechanism allows postEdit to be deployed as a post-plugin at the downstream end, safeguarding the privacy of editing data. On the other hand, an expert model called post-editor, guided by editing knowledge, makes fine-grained modifications to original responses generated by LLM, thereby effectively preserving the original style. 
As the role of post-editor is to discern and precisely edit the original response rather than storing new knowledge, we integrate editing memory and a retriever into postEdit, like IKE and SERAC, for efficient knowledge injection. 
We leave the detailed exposition in Section \ref{sec:method}.
Finally, we conduct comprehensive experiments and analysis to demonstrate that postEdit achieves outstanding performance in both editing and style retention, exhibiting robust generalization across various aspects, including LLMs, data, and scales in Section \ref{sec:experiment} and \ref{sec:analysis}.

Our contributions are three-fold: 
(1) We officially introduce knowledge editing on black-box LLMs and propose a comprehensive KE evaluation framework, incorporating the assessment of style retention for the first time.
(2) We propose a novel postEdit method to post-edit the output of LLMs through an expert model in a plug-in manner. Our postEdit can both maintain the privacy of downstream editing data and achieve commendable editing performance and style retention.
(3) Experiments and analysis on two benchmarks demonstrate that our postEdit outperforms all baselines in both editing and style retention (Retention Score $+20.82\%\uparrow$), showing robust generalization.

\begin{figure*}[t]
    \centering
    \resizebox{1\textwidth}{!}{
    \includegraphics{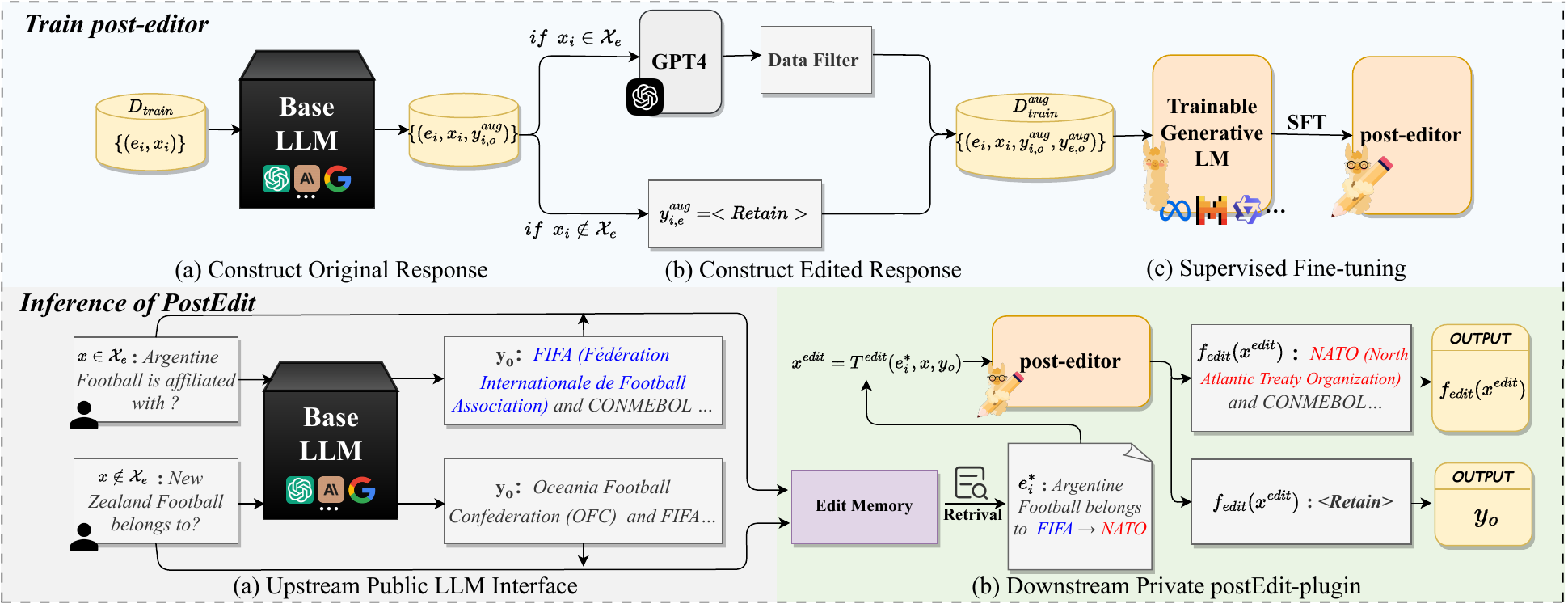}}
    \vspace{-0.8cm}
    \caption{The overall architecture of postEdit. The post-editor is trained to learn: (1) distinguish between INS and OOS queries; (2) edit the output of INS queries while preserving style. Pseudo-code is provided in Appendix \ref{subapp:postedit}.}
    \vspace{-0.2cm}
    \label{fig:method}
\end{figure*}

\section{Evaluation Framework}
\label{sec:eval}
\subsection{Problem Formulation}
 A knowledge entry is typically shown as a triple (subject, relationship, object).
Following \citet{wang2023knowledge}, an edit can be defined as $e = (t, t^{*}) = (s, r, o \to o^{*})$, denoting the update of an old knowledge triple $t$ to the new one $t^{*}$. As multiple input-output pairs can be associated with the same tuple, the input set associated with edit $e$ is denoted as $\mathcal{X}_e=I (s, r )$, referred to as in-scope (INS) input space, the target output set associated with $o^*$ is denoted as $\mathcal{Y}^{*}_{e}=O^{*}(s, r, o^{*})$, and the corresponding original output set is denoted as $\mathcal{Y}_{e} = O (s, r, o)$. For a base LLM $f_{base}: \mathcal{X}\to \mathcal{Y}$, given an edit $e$, the goal of KE is to modify the original output $y_o\in \mathcal{Y}_{e}$ to $y_e\in \mathcal{Y}_{e}^{*}$ for input $x\in \mathcal{X}_e$, while keeping the output unaffected for out-of-scope (OOS) queries, i.e., $y_e=y_o$ if $x\notin \mathcal{X}_e$. Furthermore, we define KE on black-box LLMs as the editing on a certain class of LLMs, where we have no access to anything other than textual outputs of LLMs.

\subsection{Evaluation Protocol}
\subsubsection{Existing Logit-based Evaluation}
Previous studies \cite{meng2022locating,mitchell2022memory,yao-etal-2023-editing,zheng-etal-2023-edit} primarily assess KE based on three metrics: \textbf{Efficacy}, \textbf{Generalization}, and \textbf{Specifity}, by calculating the change in logits of the model before and after editing.\footnote{We provide details of these metrics in Appendix \ref{app:metric}.} On the one hand, the inaccessibility of logits for black-box LLMs such as ChatGPT poses challenges, rendering these metrics ineffective. On the other hand, KE should only modify spans in the response involving the edit, while keeping the response's style unchanged to minimize negative impacts of editing. However, this aspect has been fully overlooked, leading to incomplete evaluation.

\subsubsection{Improved Multi-perspective Evaluation}
\label{subsec:eval_framework}
For black-box LLMs editing, the evaluation of KE focuses on what changes and what remains in the edited output $y_e$ compared to original output $y_o$. Therefore, we formulate the evaluation framework from both the aspects of editing and retention.\footnote{We prove the proposed evaluation's validity through its consistency with human evaluation in Appendix \ref{subapp:consistency}.}\\
\textbf{Editing} The Editing metric is designed to evaluate the editing for INS input and non-editing for OOS input. When $x\in \mathcal{X}_{e}$, the expected output space of $f_{base}$ transitions from $\mathcal{Y}_e$ to $\mathcal{Y}_e^*$. From the perspective of textual editing (\textbf{TE}),  $\mathcal{Y}_{e}^{*}$ discards the old target $o$ and incorporates the new target $o^*$. From the perspective of semantic editing (\textbf{SE}), the joint text composed of $\mathcal{X}_{e}$ and $\mathcal{Y}_{e}^{*}$ implies the new knowledge $t^*$ and contradicts the old knowledge $t$. When $x\notin \mathcal{X}_{e}$, the situation is reversed. We formalize TE as follows:
\begin{equation}
\resizebox{1\hsize}{!}{$
\operatorname{TE}=\begin{cases}\frac{1}{2}\{\operatorname{ctn}(y_e,o^*)+(1-\operatorname{ctn}(y_e,o))\}\ x\in \mathcal{X}_e\\\frac{1}{2}\{\operatorname{ctn}(y_e,o)+(1-\operatorname{ctn}(y_e,o^*))\}\ x \notin \mathcal{X}_e  \end{cases}
$}
\end{equation}
where $\operatorname{ctn}(a,b)=1$ if a \textbf{contains} b, otherwise 0. 
Similarly, SE is formalized as follows:
\begin{equation}
\resizebox{1\hsize}{!}{$
\operatorname{SE}=\begin{cases}\frac{1}{2}\{\operatorname{ent}([x,y_e],t^*)+(1-\operatorname{ent}([x,y_e],t))\} \ x\in \mathcal{X}_e\\\frac{1}{2}\{\operatorname{ent}([x,y_e],t_{o})+(1-\operatorname{ent}([x,y_e],t^*))\} \ x\notin \mathcal{X}_e  \end{cases} 
$}
\end{equation}
where $\operatorname{ent}(a, b) = 1$ if a \textbf{entails} b, otherwise 0 by using the Natural Language Inference (NLI) model, $[x, y_e]$ denotes the concatenation of input-output pair , and $t_{o}$ indicates the knowledge tuple associated with OOS input-output pair $[x, y_o]$.\\
\textbf{Retention} To assess the extent to which the edited output preserves the original style, we introduce Retention as an adversarial metric for Editing. We separately evaluate textual retention (\textbf{TR}) and semantic retention (\textbf{SR}) using ROUGE scores \cite{lin2004rouge} and the SBERT model \cite{reimers-gurevych-2019-sentence}, formalized as follows:
\begin{equation}
\operatorname{TR}=\begin{cases}\operatorname{ROUGE}(\operatorname{M}(y_e,o^*),\operatorname{M}(y_o,o)) & x\in \mathcal{X}_e\\\operatorname{ROUGE}(y_e,y_o) & x\notin \mathcal{X}_e \end{cases} 
\end{equation}
\begin{equation}
\operatorname{SR}=\begin{cases}\operatorname{sim}(\operatorname{M}(y_e,o^*),\operatorname{M}(y_o,o)) &\ x\in \mathcal{X}_e\\\operatorname{sim}(y_e,y_o) &\ x\notin \mathcal{X}_e \end{cases} 
\end{equation}
where $\operatorname{M}(a, b)$ denotes masking the span relevant to b in a. For $x \in \mathcal{X}_e$, we employ a masking operation to extract text unrelated to editing\footnote{We show pseudo-code of these metrics in Appendix \ref{subapp:evalutaion_detail}.}.

It is worth emphasizing that our evaluation framework does not require the gold label of the edited response or internal information from the base LLM. This enables its applicability to a wide range of scenarios beyond black-box LLM editing.

\section{Methodology}
\label{sec:method}

\subsection{Overall Architecture}
\label{subsec:architecture}
To solve the problems of privacy leakage of editing data and style over-editing, as illustrated in Fig \ref{fig:method}, postEdit is deployed downstream and post-processes the output of base LLM, comprising three components: an edit-memory $M_e=\{e_i\}$ for storing editing knowledge, a retriever $f_{retr}$ for recalling an edit, and a trained generative model named post-editor $f_{edit}$ for executing the edit\footnote{In the main experiment, we fine-tune LLaMA 2-7B \cite{touvron2023LLaMA} as the post-editor and conduct an analysis of performance at various scales in Section \ref{subsec:scale}.
}. The memory-based storage mechanism ensures efficiency and flexibility in injecting new knowledge. During the inference phase, the retriever first recalls the edit with the highest similarity to user input from $M_e$. Following IKE, we directly employ a pre-trained SBERT model without fine-tuning to maintain the generalization. Finally, the post-editor performs the editing 
guided by recalled edit.

\subsection{Train post-editor}
\textbf{Original Response Augment} The training dataset of KE typically consists of editing knowledge, along with queries covering both INS and OOS input, denoted as $D_{train}=\{(e_i,x_i)\}$. Previous studies \cite{mitchell2022memory,zheng-etal-2023-edit} usually directly use the new object $o^*_i$ in $e_i$ as the target output for editing, resulting in stylistic differences between the editor and base LLM. To address this gap, we first construct the original response $y^{aug}_{i,o}=f_{base}(x_i)$ via base LLM for each sample.\\
\textbf{Edited Response Augmentation} In order to construct the training output targets for post-editor, we utilize both GPT-4 and rules to further augment the training dataset.  For INS inputs, the objective is to modify the original response. Thus, given edit $e_i$, input $x_i$, and original output $y^{aug}_{i,o}$ are aggregated using an editing template $T^{aug}$\footnote{All templates mentioned are shown in Appendix \ref{subapp:prompt}.} and fed into GPT-4 to obtain the edited output $y^{aug}_{i,e}$. For OOS inputs, the goal is to maintain the original response without modification. Therefore, we introduce a special token $\left<Retain\right>$ as the target output, denoting no need for editing. We formulate this process as:
\begin{equation}
y^{aug}_{i,e}=\begin{cases}f_{gpt4}(T^{aug}(e_i,x_i,y^{aug}_{i,o}))  &  x_i \in \mathcal{X}_e\\ \left<Retain\right> & x_i \notin \mathcal{X}_e\end{cases}
\end{equation}
Recent studies \cite{zhou2023lima,lu2023instag,liu2023makes} have proven that the quality of training data is often more crucial than quantity. To further enhance the quality of augmented data and alleviate training burden, we evaluate and filter the edited responses obtained through GPT-4 augment. Based on the joint evaluation using the Editing metrics TE and SE, formalized as $\mathbf{1}_{\{\operatorname{TE}=1\&\operatorname{SE}=1\}}y^{aug}_{i,e}$, augmented samples with poor quality are discarded. Ultimately, we obtain the augmented training set $D^{aug}_{train}=\{(e_i,x_i,y^{aug}_{i,o},y^{aug}_{i,e})\}$.\\
\textbf{Supervised Fine-tuning (SFT)} After data augment and filtering, the post-editor is trained in a supervised fine-tuning manner, where the query, edit, and original response are aggregated as input using an editing template $T^{edit}$ (distinct from $T^{aug}$), with $y^{aug}_{i,e}$ as the output target. After tokenizing $y^{aug}_{i,e}$ as $\{y^{aug}_{i,e_1},y^{aug}_{i,e_2},…,y^{aug}_{i,e_T}\}$, the loss function of SFT can be formalized as follows:
\begin{equation}
\label{eq:sft}
\mathcal{L}_{sft}=-\sum_{i=1}^{|D^{aug}_{train}|} \sum^{T-1}_{t=0} log \\ P(y^{aug}_{i,e_{t+1}}|x^{edit}_i,y_{i,e_{\leq t}})
\end{equation}
where $x^{edit}_i=T^{edit}(e_i,x_i,y^{aug}_{i,o})$.

\begin{table*}[t]
    \setlength\tabcolsep{4pt}
    \centering
     \resizebox{1\textwidth}{!}{%
    \begin{tabular}{c|cccc|cccc||cccc|cccc}
    \hline
        \multirow{2}{*}{Method} & \multicolumn{4}{c|}{Textual Editing (TE)} & \multicolumn{4}{c||}{Semantic Editing (SE)} & \multicolumn{4}{c|}{Textual Retention (TR)} & \multicolumn{4}{c}{Semantic Retention (SR)}\\ \cline{2-17}
         ~ & Simple & Rephrase & OOS & AVG & Simple & Rephrase & OOS & AVG & Simple & Rephrase & OOS & AVG & Simple & Rephrase & OOS & AVG\\ \hline
        PROMPT & 85.17 & 86.73 & 63.8 & 78.57 & 83.1 & 84.57 & 61.97 & 76.54  & 21.42 & 21.54 & 18.11 & 20.36 & 53.14 & 54.86 & 51.37 & 53.13\\ 
        IKE & 94.2 & 85.8 & 85.4 & 88.47 & 93.2 & 84.5 & 85.3 & 87.67  & 24.14 & 18.98 & 22.81 & 21.97 & 53.45 & 48.94 & 57.69 & 53.36\\ \hline
        SERAC & \underline{95.4} & \underline{87.4} & 96.1 & 92.97 & \underline{94.6} & \underline{87.3} & 96.2 & 92.7 & \underline{35.66} & \underline{37.62} & 96.01 & \underline{56.43} & \underline{65.51} & \underline{64.64} & 97.04 & \underline{75.73}\\
        SERAC (ChatGPT) & 95.23 & 85.8 & \underline{98.6} & \underline{93.2} & \textbf{95.3} & 86 & \underline{98.6} & \underline{93.31} & 23.43 & 26.71 & \underline{96.41} & 48.85 & 55.04 & 56.88 & \underline{97.91} & 69.95\\ \hline 
        postEdit (ours) & \textbf{96.8} & \textbf{94.7} & \textbf{99.4} & \textbf{96.97} & 92.5 & \textbf{92.1} & \textbf{99.4} & \textbf{94.67} & \textbf{88.65} & \textbf{89.66} & \textbf{99.64} & \textbf{92.65} & \textbf{93.9} & \textbf{94.02} & \textbf{99.82} & \textbf{95.91}\\ \hline
\end{tabular}
}
\vspace{-0.3cm}
\caption{Performance comparison on CounterFact. We bold the best results and underline the second-best results. Results are averaged over three random runs (p < 0.01 under t-test).}
\vspace{-0.2cm}
\label{tab:main_cf}
\end{table*}

\begin{table*}[t]
    \centering
    \setlength\tabcolsep{4pt}
     \resizebox{1\textwidth}{!}{%
    \begin{tabular}{c|cccc|cccc||cccc|cccc}
    \hline
        \multirow{2}{*}{Method} & \multicolumn{4}{c|}{Textual Editing (TE)} & \multicolumn{4}{c||}{Semantic Editing (SE)} & \multicolumn{4}{c|}{Textual Retention (TR)} & \multicolumn{4}{c}{Semantic Retention (SR)}\\ \cline{2-17}
         ~ & Simple & Rephrase & OOS & AVG & Simple & Rephrase & OOS & AVG & Simple & Rephrase & OOS & AVG & Simple & Rephrase & OOS & AVG\\ \hline
        PROMPT & 88.83 & 86.87 & 58.37 & 78.02 & 86.5 & 84.97 & 60.27 & 77.24 & 47.76 & 45.35 & 34.93 & 42.68 & 73.4 & 74.62 & 61.29 & 69.77\\ 
        IKE & 98.1 & 97.6 & 78 & 91.23 & \textbf{97.7} & \underline{94.7} & 83.1 & 91.83 & 19.72 & 16.36 & 27.83 & 21.3 & 42.26 & 38.67 & 58.53 & 46.49 \\ \hline
        SERAC & \textbf{98.7} & 95.1 & \textbf{100} & \underline{97.93} & \underline{97.6} & 93.3 & \textbf{100} & \underline{96.97} & \underline{68.02} & \underline{66.06} & \textbf{100} & \underline{78.03} & \underline{86.84} & \underline{85.91} & \textbf{100} & \underline{90.92} \\ 
        SERAC (ChatGPT) &  94.7 & \underline{87.5} & \textbf{100} & 94.07 & 96.17 & 88.53 & \textbf{100} & 94.9 & 52.22 & 52.01 & \textbf{100} & 68.08 & 75.2 & 77.56 & \textbf{100} & 84.25\\ \hline
        postEdit (ours) & \underline{98.4} & \textbf{98.6} & \textbf{100} & \textbf{99} & 96.2 & \textbf{95.4} & \textbf{100} & \textbf{97.2} & \textbf{95.76} & \textbf{96.13} & \textbf{100} & \textbf{97.3} & \textbf{97.69} & \textbf{97.89} & \textbf{100} & \textbf{98.53} \\ \hline
\end{tabular}
}
\vspace{-0.3cm}
\caption{Performance comparison on zsRE.}
\vspace{-0.2cm}
\label{tab:main_zsre}
\end{table*}

\subsection{Inference of PostEdit}
For a user query $x\in D_{test}$, the original response $y_o=f_{base}(x)$ is obtained through the upstream LLM interface. On the downstream side, the retriever recalls the most similar edit $e_{i^*}$ to $x$ from the edit memory:
\begin{equation}
i^*=\operatorname{argmax}_{0\leq i<|M_e|}\ \operatorname{sim}(x,e_i)
\end{equation}
Next, we obtain the input $x^{edit}=T^{edit}(e_{i^*},x,y_o)$ by populating the editing template $T^{edit}$ and transmit it to the post-editor to yield the output $f_{edit}(x^{edit})$. Finally, by discerning whether $f(x^{edit})$ contains the special token $\left<Retain\right>$, we determine the ultimate output:
\begin{equation}
    y_e=\begin{cases}f_{edit}(x^{edit})  &  f_{edit}(x^{edit})\neq \left<Retain\right>\\ y_o & f_{edit}(x^{edit})= \left<Retain\right>\end{cases}
\end{equation}


\section{Experiments}
\label{sec:experiment}

\subsection{Datasets}
We conduct experiments on two widely-used KE datasets, CounterFact \cite{meng2022locating} and zsRE \cite{levy2017zero}. Each entry comprises an edit and three types of queries:  \textbf{Simple} queries to validate the success of knowledge injection, \textbf{Rephrase} queries to assess the generalization of the edit, and \textbf{out-of-scope (OOS)} queries to verify the local effect of the edit. Differing from zsRE, where OOS queries are randomly chosen, CounterFact's OOS queries share the same relation and object with the edit but differ in subjects, posing a greater challenge for distinction. We provide details and processing procedures in Appendix \ref{subapp:dataset}.

\subsection{Baselines}
We employ ChatGPT as the base LLM and extensively compare postEdit with methods applicable to black-box LLM editing, including PROMPT \cite{zheng-etal-2023-edit}, IKE \cite{zheng-etal-2023-edit}, SERAC \cite{mitchell2022memory}, and SERAC(ChatGPT). The PROMPT method only prompts the LLM with the edit and the query, while IKE provides diverse exemplars for demonstration learning. SERAC employs a fine-tuned surrogate model\footnote{For a fair comparison, the surrogate model uses the same pre-trained model and training data as the post-editor.} to respond to queries within the editing scope, and SERAC(ChatGPT) is a variant where the surrogate model is changed to ChatGPT. More details are shown in Appendix \ref{subapp:baselines}.

\begin{table*}[t]
    \centering
    \resizebox{0.98\textwidth}{!}{%
    \begin{tabular}{c|c|c|c|c|c|c}
    \hline
    \multirow{2}{*}{\thead{ID}} & \multirow{2}{*}{\thead{Edit}} & \multirow{2}{*}{\thead{Query}} & \multirow{2}{*}{\thead{Original Response}} & \multicolumn{3}{c}{\thead{Edited Response}}\\ \cline{5-7}
    ~ &~ &~ &~ &\thead{IKE} &\thead{SERAC} &\thead{postEdit} \\ \hline
    \thead{1} & \thead{The nationality of Marcel\\Maupi was what? \textcolor{blue}{French} $\rightarrow$ \textcolor{red}{Italian}} & \thead{What was Marcel\\Maupi's nationality?} & \thead{Marcel Maupi was a\\ \textcolor{blue}{French} national.}& \thead{Italian.} &\thead{Marcel Maupi\\was Italian.} &\thead{Marcel Maupi was an\\\textcolor{red}{Italian} national.}\\ \hline
    \thead{2} & \thead{Football Association of\\Malaysia is affiliated\\ with the \textcolor{blue}{FIFA} $\rightarrow$ \textcolor{red}{Avengers}} & \thead{Football Association \\of Malaysia is\\affiliated with \_} & \thead{- Asian Football Confederation (AFC)\\- \textcolor{blue}{FIFA (Fédération Internationale}\\\textcolor{blue}{de Football Association)}}& \thead{Avengers.} &\thead{Avengers.} &\thead{- Asian Football Confederation (AFC)\\- \textcolor{red}{Avengers}}\\ \hline
    \thead{3} & \thead{What is the programming\\language for CLPython?\\ \textcolor{blue}{Lisp} $\rightarrow$ \textcolor{red}{JavaScript}} & \thead{What is the\\programming language\\for CLPython?} & \thead{CLPython is an implementation\\of Python language in \\\textcolor{blue}{Common Lisp}, which means\\that the programming language\\used for CLPython is \textcolor{blue}{Common Lisp}.}& \thead{JavaScript.} &\thead{The programming\\language for\\CLPython is\\JavaScript.} &\thead{CLPython is an implementation\\of Python language in\\\textcolor{red}{JavaScript},which means\\that the programming language\\used for CLPython is \textcolor{red}{JavaScript}.}\\ \hline
    \thead{4} & \thead{Howard Glacier is located\\in \textcolor{blue}{Antarctica} $\rightarrow$ \textcolor{red}{Europe}} & \thead{Howard Glacier\\is located in\\the continent \_} & \thead{of \textcolor{blue}{Antarctica}. Specifically, it is \\located in the \textcolor{blue}{northern part}\\ \textcolor{blue}{of the Antarctic Peninsula}.}& \thead{Europe.} &\thead{of Europe.} &\thead{of \textcolor{red}{Europe}. Specifically, it is\\located in the \textcolor{red}{northern part}\\\textcolor{red}{of the continent}.}\\ \hline
    \thead{5} & \thead{In what city or state did\\the formation of I Am Kloot\\occur? \textcolor{blue}{Manchester} $\rightarrow$ \textcolor{red}{Bilbao}} & \thead{In which city or state\\was the formation\\of I Am Kloot?} & \thead{I Am Kloot was formed\\in \textcolor{blue}{Manchester, England}.}& \thead{Bilbao.} &\thead{I Am Kloot was\\formed in Bilbao.} &\thead{I Am Kloot was formed\\in \textcolor{red}{Bilbao, Spain}.}\\ \hline
    \end{tabular}
    }
\vspace{-0.2cm}
\caption{Editing cases sampled from CounterFact and zsRE under different methods.}
\vspace{-0.1cm}
\label{tab:case}
\end{table*}

\subsection{Main Results}
Table \ref{tab:main_cf} and Table \ref{tab:main_zsre} show the main results of postEdit and comparable baselines on two benchmark KE datasets. In general, our postEdit method consistently outperforms all baselines with a large margin, both in terms of Editing and Retention scores. Next, we analyze the results from three aspects:

(1) \textbf{Comparison of different methods.} We can see that postEdit achieves nearly all optimal Editing scores, along with a significant surpassing of baselines in Retention scores. On CounterFact, postEdit outperforms the suboptimal baselines by 3.77\% (TE), 1.36\% (SE), 36.22\% (TR), and 20.18\% (SR) in average scores. On zsRE, postEdit surpasses the suboptimal baselines by 1.07\% (TE), 0.23\% (SE), 19.27\% (TR), and 7.61\% (SR). This shows that postEdit can accurately locates and modifies spans in the text related to editing, while maintaining other content, thereby achieving high performance in both Editing and Retention.

(2) \textbf{Comparison of different query types.} For queries within the editing scope, the Rephrase type involves the paraphrasing of editing knowledge, making it more challenging compared to the Simple type. Concerning CounterFact, discernible decrements in Rephrase performance are observed for IKE and SERAC in contrast to the Simple type (e.g., TE score, IKE: 94.2→85.8, SERAC: 95.5→87.4), whereas postEdit performance remains stable (96.8→94.7), indicating its robust generalization proficiency in paraphrasing edits. For OOS queries, while SERAC and postEdit excel on the zsRE dataset, postEdit  surpasses  SERAC on more challenging CounterFact, showcasing its precise differentiation of queries requiring editing without additional editing judge module.

(3) \textbf{Comparison of different metrics.} Comparing the Editing and Retention of baselines reveals a serious issue of style over-editing. For example, the Editing performance of IKE surpasses that of PROMPT, while the Retention lags behind PROMPT, indicating a negative impact of demonstration on IKE's style retention. Despite achieving commendable Editing scores, SERAC and SERAC (ChatGPT) still fall short in terms of Retention. This highlights that effective editing does not guarantee good retention, emphasizing the need for a comprehensive evaluation of knowledge editing.

\section{Analysis}
\label{sec:analysis}

\subsection{Generalization of PostEdit}
\begin{figure}[t]
    \centering
    \resizebox{0.48\textwidth}{!}{
    \includegraphics{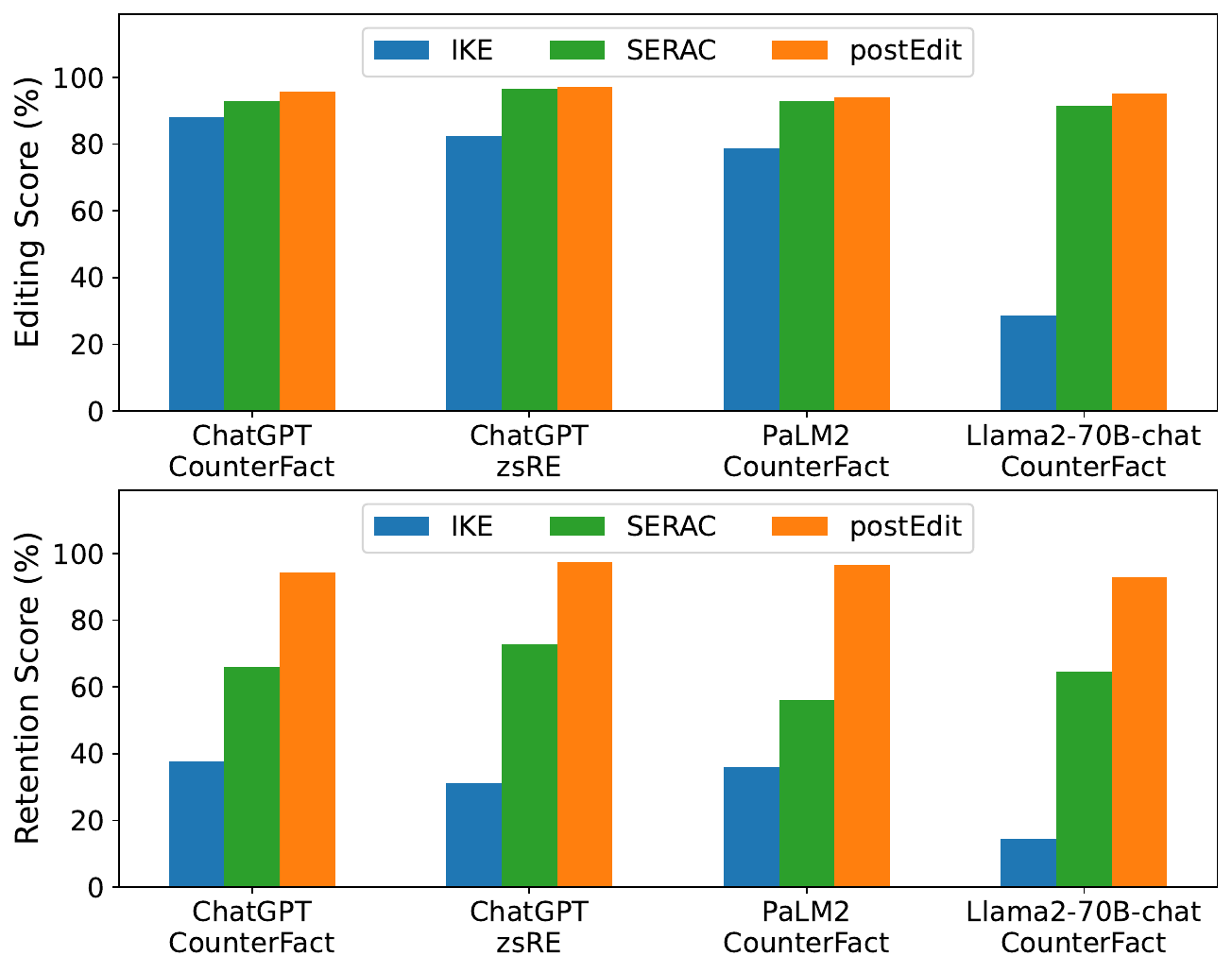}}
    \vspace{-0.8cm}
    \caption{Performance under different 
    base LLMs and datasets, where Editing Score is the average of TE and SE, and Retention Score is the average of TR and SR.}
    \vspace{-0.4cm}
    \label{fig:generalization}
\end{figure}

In Section \ref{subsec:architecture}, we fine-tune the post-editor to acquire the ability of discriminating and executing edits. Therefore, it is imperative to validate the generalization of post-editor's abilities. For postEdit and baselines, we initially utilize ChatGPT as the base LLM and CounterFact as the training set or demonstration library. Subsequently, we conduct testing under different base LLMs and datasets without re-training, as illustrated in Fig \ref{fig:generalization}. 

We can see that whether generalizing from CounterFact to zsRE or from ChatGPT to PaLM2\footnote{https://ai.google/discover/palm2} and LLaMA2-70B-chat\footnote{https://huggingface.co/meta-LLaMA}, postEdit consistently demonstrates optimal performance in Editing and Retention scores. This substantiates the robust generalization of postEdit, highlighting its plug-and-play applicability across diverse scenarios without retraining. In contrast, both IKE and SERAC exhibit performance fluctuations, particularly evident in a significant decline when IKE is applied to LLaMA2-70B-chat. Further analysis reveals that conflicts between editing data and the intrinsic knowledge of LLaMA2-70B-chat lead to frequent refusals to generate responses based on edits. However, postEdit successfully mitigated the impact of knowledge conflicts through post-processing.

\subsection{Case Study}
To visually demonstrate the editing and style retention of postEdit and baselines, we conduct the case study  in Table \ref{tab:case}.
In Case 1, postEdit accurately identifies and modifies "\emph{French}" to "\emph{Italian}" while maintaining the rest of the text unchanged to keep the style to the greatest extent. In contrast, IKE only responds with "\emph{Italian}" and SERAC replies with "\emph{Marcel Maupi was Italian}" without referencing the original response, revealing serious style over-editing. 
In Cases 2 and 3, postEdit respectively replaces "\emph{FIFA (Fédération Internationale de Football Association)}" with "\emph{Avengers}" and modifies "\emph{Common Lisp}" to "\emph{JavaScript}". This demonstrates that postEdit can locate and edit spans semantically related to editing knowledge, going beyond a rudimentary replacement of old objects with new ones.
Furthermore, it is evident that postEdit can handle spans logically associated with the editing. In Case 4, the location changes from "\emph{Antarctica}" to "\emph{Europe}", and the span in the original response, describing the location as "\emph{the northern part of the Antarctic Peninsula}", is correspondingly adjusted to "\emph{the northern part of the continent}". Similarly, in Case 5, as "\emph{Manchester}" is changed to "\emph{Bilbao}", the country is also edited from "\emph{England}" to "\emph{Spain}".

\begin{table}[t]
    \centering
    \setlength\tabcolsep{4pt}
     \resizebox{0.48\textwidth}{!}{%
    \begin{tabular}{c|cccc|cccc}
    \hline
        \multirow{2}{*}{Method} & \multicolumn{4}{c|}{Semantic Editing (SE)} & \multicolumn{4}{c}{Semantic Retention (SR)} \\ \cline{2-9}
         ~ & Simple& Rephrase & OOS & AVG & Simple & Rephrase & OOS & AVG \\ \hline
         postEdit & 92.5 & 92.1 & 99.4 & \textbf{94.67}  & 93.9 & 94.02 & 99.82 & \textbf{95.91} \\ \hline
         \multicolumn{9}{c}{\emph{Module Ablation}}  \\ \hline
         -w/o data fillter & 90.6 & 90.6 & 99.4 & 93.53  & 94.19 & 93.76 & 99.82 & 95.92 \\ 
         post-editor$\rightarrow$ChatGPT & 89.73 & 87.8 & 70.77 & 82.54  & 89.39 & 88.78 & 83.27 & 86.26\\
         GPT4$\rightarrow$ChatGPT & 93.2 & 91.8 & 99.4 & 94.80  & 90.04 & 89.54 & 99.81 & 93.13\\
         SBERT Judgement & 92.2 & 85.2 & 96.3 & 91.23  & 94.47 & 92.49 & 98.97 & 95.31 \\ \hline
         \multicolumn{9}{c}{\emph{Training Data Ablation}}  \\ \hline
        -w/o Simple & 91.8 & 91.2 & 99.5 & 94.17 & 93.96 & 94.21 & 99.89 & 96.02 \\ 
        -w/o Rephrase & 92 & 12.9 & 99.8 & 68.23 & 94.37 & 71.67 & 99.95 & 88.66 \\ 
        -w/o OOS & 92.2 & 91.5 & 4.7 & 62.8 & 94.47 & 94.12 & 75.01 & 87.86 \\ \hline
\end{tabular}
}
\vspace{-0.3cm}
\caption{Ablation Study on CounterFact.}
\vspace{-0.5cm}
\label{tab:ablation}
\end{table}

\subsection{Ablation Study}
\label{subsec:ablation}
To understand the roles of each component and training data type in postEdit, we conduct ablation study in Table \ref{tab:ablation}.\\
\textbf{Module Ablation} In our postEdit framework, we utilize GPT-4 to generate edited responses and subsequently perform data filtering. 
After removing data filtering, the SE score for INS queries exhibits a decline (Simple -1.9 and Rephrase -1.5), indicating that data filtering effectively enhances the quality of training data. 
Replacing the post-editor with ChatGPT results in a noticeable decline in performance across different types. This suggests that LLMs like ChatGPT are not proficient performing such editing tasks, highlighting the need for fine-tuning the post-editor.
Substituting GPT-4 with ChatGPT for edited response augmentation results in a slight SE score increase (avg +0.13) but a significant SR score decrease (avg -2.78). This indicates that ChatGPT lacks the fine-grained granularity in editing compared to GPT-4, thereby resulting in a coarser-grained post-editor.
Finally, we introduce the editing judging module, the same as SERAC, through comparing the SBERT semantic similarity with a threshold. The observed decrease in Rephrase and OOS scores demonstrates the superior discriminative capability of the post-editor.\\
\textbf{Training Data Ablation} We further conduct data ablation by removing each type of data from the training set. We observe that removing Simple data has no notable impact, while the removal of Rephrase data leads to a significant drop (-79.2) in the SE metric. This indicates that Rephrase data plays a crucial role in improving the post-editor's ability for editing knowledge injection and generalization, while relying solely on Simple data doesn't suffice for achieving the post-editor's generalization.
After removing OOS data, although there is a noticeable decline in OOS metrics, the metrics for Simple and Rephrase do not show a discernible improvement. This indicates that post-editor doesn't excessively compromise its ability to perform edits when learning to discriminate editing.

\subsection{Effect of Post-editor Scale}
\label{subsec:scale}
\begin{figure}[t]
    \centering
    \resizebox{0.48\textwidth}{!}{
    \includegraphics{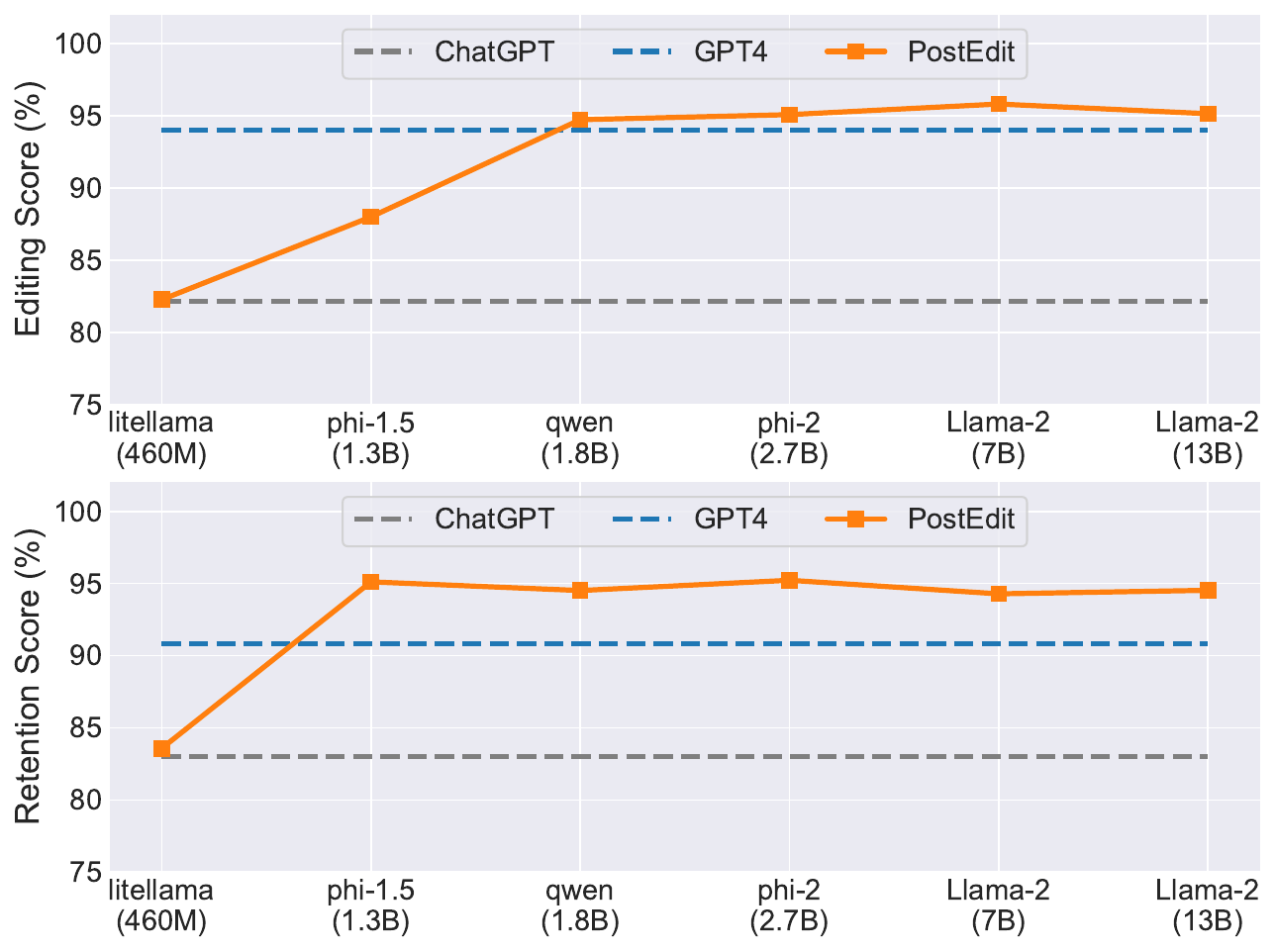}}
    \vspace{-0.7cm}
    \caption{Performance curves of the post-editor at different scales on CounterFact.}
    \vspace{-0.5cm}
    \label{fig:scaling}
\end{figure}
To investigate the effect of post-editor scale on performance, we compare evaluation scores across models ranging from 460M to 13B in size. As illustrated in Fig \ref{fig:scaling}, it is evident that with the increase in post-editor scale, editing scores gradually improve (significant from 460M to 1.8B, followed by slower gains beyond 1.8B), while retention score remains stable after reaching 1.3B. This suggests that editing ability is more influenced by the model scale, and a larger post-editor can enhance editing performance while maintaining the retention. We also compare the effectiveness of post-editor with zero-shot ChatGPT and GPT-4. Similar to the findings in Section \ref{subsec:ablation}, LLMs like ChatGPT are not proficient in executing the editing task. Therefore, on CounterFact, the performance of the 460M post-editor is comparable to ChatGPT, and the 1.8B post-editor surpasses GPT-4. This indicates that the postEdit framework does not rely on a large-scale post-editor, and small-sized editors can achieve satisfactory performance and high efficiency.

\subsection{Effect of Memory Size}
\begin{figure}[t]
    \centering
    \resizebox{0.48\textwidth}{!}{
    \includegraphics{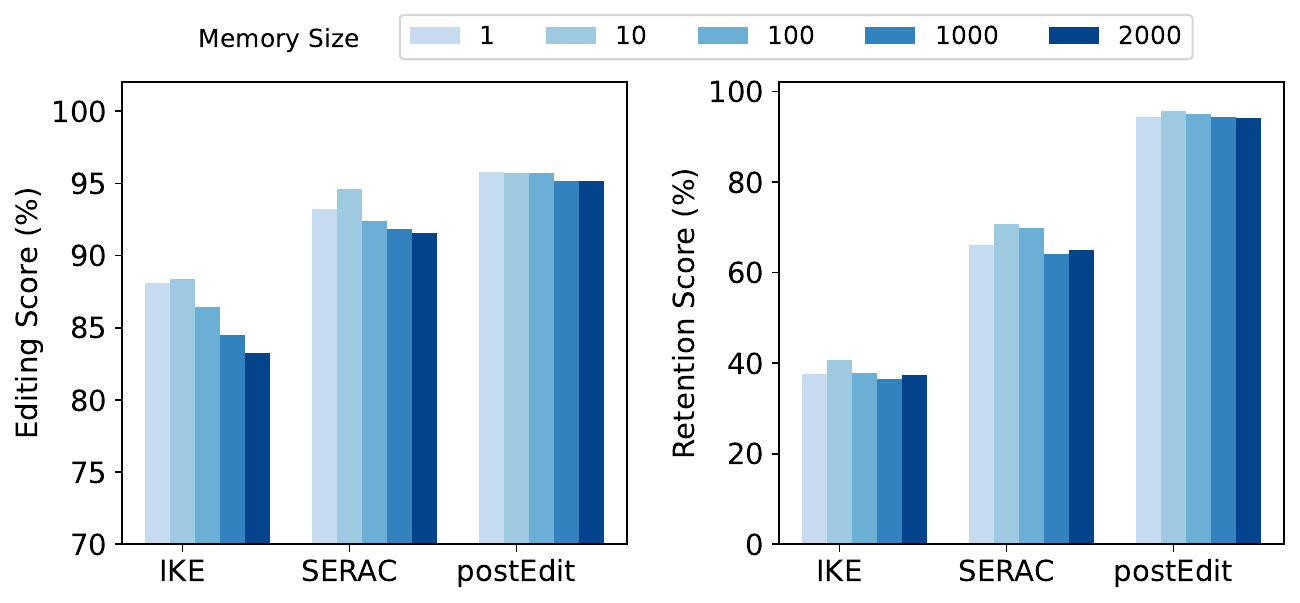}}
    \vspace{-0.7cm}
    \caption{Performance of methods under different Editing Memory size on CounterFact.}
    \vspace{-0.4cm}
    \label{fig:memory}
\end{figure}
It should be noted that the default evaluation procedure of KE is to edit a single piece of knowledge, assess, and then roll back the editing before repeating the process for each test point. However, in real-world scenarios, as the world evolves, edited knowledge should be continuously infused and preserved, i.e., the size of Editing Memory will continue to expand\footnote{In some studies, this corresponds to Batch Editing and Sequence Editing.}. 
For the edit retrieved from Editing Memory, IKE utilizes the base LLM itself, SERAC applies a similarity threshold, and postEdit employs the post-editor to determine whether the query is within the scope of editing.
We evaluate the performance of these methods under varying memory sizes in Fig \ref{fig:memory}. With the same retriever,  postEdit exhibits the highest robustness among methods in both Editing and Retention scores, substantiating the superiority of the postEdit mechanism in discerning the necessity of editing.

\subsection{Discussion on Efficiency} 
Apart from Editing and Retention performance, KE methods should strive to minimize storage and computational costs. For memory-based black-box LLM editing, 
in addition to Editing Memory and the retriever, storage overhead also encompasses the demonstration library for IKE, the judge model and surrogate model for SERAC, and the post-editor for postEdit.
Furthermore, although memory-based methods do not incur computational overhead for editing,they do introduce inference expenses. Specifically, for IKE, the inference cost increases from $f_{base}(x)$ to $f_{retr}(x,M_e)+{f_{base}(demos,e,x)}$; for SERAC, the additional cost is $f_{retr}(x,M_e)+f_{judge}(x,e_{retr})$; and for postEdit, it is $f_{retr}(x,M_e)+f_{edit}(e,x,y_o)$.
To further reduce post-editing overhead, one approach is to improve the reasoning efficiency of the post-editor. As highlighted in Section \ref{subsec:scale}, a small-scale post-editor can also achieve commendable performance. Another potential option is to employ white-box parameter-editing methods to directly integrate editing knowledge into the post-editor. The post-editor can then use its knowledge to modify the original response of base LLM, exchanging editing costs for memory storage and retrieval expenses.

\section{Related Work}
\label{sec:relate}
\textbf{Knowledge Editing} 
The initial methods for knowledge editing involve updating parameters using constrained fine-tuning  \cite{sinitsin2020editable,zhu2020modifying}. 
Recent studies mostly center around hyper-network and attribution. Hyper-network-based approaches \cite{de-cao-etal-2021-editing,mitchell2021fast} train a hyper-network to capture gradient changes for specific edits, while attribute-based methods \cite{dai-etal-2022-knowledge,meng2022locating,meng2022mass,li2023pmet,wu-etal-2023-depn} locate neuron activation in 
 networks for targeted parameter updates. However, these approaches exclusively focus on editing in white-box LLM scenarios, overlooking concerns related to editing data privacy and stylistic consistency. Consequently, we propose a novel evaluation framework and postEdit method for black-box LLM editing to address these issues.\\
\textbf{Post-processing Methods} 
\citet{cao-etal-2020-factual} fine-tune a BART model to improve factual consistency in abstractive summarization by using summaries with errors as input and original or gold summaries as training targets. \citet{thorne-vlachos-2021-evidence} fine-tune a T5 model to correct factual errors by recovering masked statements based on retrieved evidence. RARR \cite{gao-etal-2023-rarr} 
employs PaLM with few-shot demonstrations for error correction and attribution report generation. Different from these studies, postEdit applies post-processing to the knowledge editing task, fine-tuning the post-editor to simultaneously determine query relevance within the editing scope and make fine-grained modifications.

\section{Conclusion}
\label{sec:conclusion}
In this paper, we firstly introduce a comprehensive evaluation framework for knowledge editing under black-box LLMs, incorporating multiple perspectives and considering the style retention. Next, we propose a novel postEdit framework to address existing issues in privacy leakage of editing data and style over-editing  in current methods by post-processing the output of LLMs. Finally, experiments on two benchmarks and thorough analysis demonstrate that postEdit outperforms all baselines and achieves strong generalization.

\section*{Limitations}
This paper primarily investigates the assessment and methodology of knowledge editing in black-box LLM scenarios. The proposed evaluation framework can comprehensively assess edited responses from multiple perspectives, and the postEdit method effectively addresses issues related to privacy concerns of editing data and style over-editing. However, our work also has several limitations: 
(1) Although our proposed evaluation framework and postEdit method mainly focus on knowledge editing in black-box LLM scenarios, they can be equally applied to editing in white-box LLM scenarios. Due to constraints in length and the focus of the paper, we haven't thoroughly explored this in the paper.
(2) Although the postEdit framework does not require retraining when injecting editing knowledge, it still necessitates an initial fine-tuning phase to enable the post-editor to learn the ability to discern whether a query is within the editing scope and how to perform the editing, resulting in a certain computational load.
(3) Our study primarily investigates the application of knowledge editing in knowledge question answering tasks, similar to previous research.  We believe that our framework can be extended to other scenarios, such as fact-checking and sentiment editing. We leave these explorations for future research.

\section*{Ethic Consideration}
In this paper, we propose a knowledge editing approach that can be flexibly applied downstream  to post-process the outputs of LLMs, effectively safeguarding the privacy of downstream private editing data and maintaining consistency in the style of the LLM. 
While the purpose of knowledge editing is to rectify errors or outdated knowledge in LLMs, malicious knowledge editing may lead to the generation of harmful or inappropriate outputs by the model. Therefore, ensuring secure and responsible practices in knowledge editing is of paramount importance. The application of these techniques should be guided by ethical considerations, with safeguard measures in place to prevent misuse and mitigate the potential for harmful outcomes. 
Additionally, due to the difficulty in obtaining continuously up-to-date knowledge, some KE datasets such as CounterFact use counterfactual knowledge to validate the effectiveness of methods.  
Furthermore, the base LLM, such as ChatGPT used in this work, merely serves as a demonstration of research on knowledge editing in black-box model scenarios.
We emphasize that these datasets and LLMs are solely for academic exploration and do not involve actual applications in real-world scenarios, nor do they include content modification or attacks on commercially used LLMs.
\bibliography{custom}

\begin{thebibliography}{33}
\expandafter\ifx\csname natexlab\endcsname\relax\def\natexlab#1{#1}\fi

\bibitem[{Bian et~al.(2023)Bian, Han, Sun, Lin, Lu, and He}]{bian2023chatgpt}
Ning Bian, Xianpei Han, Le~Sun, Hongyu Lin, Yaojie Lu, and Ben He. 2023.
\newblock Chatgpt is a knowledgeable but inexperienced solver: An investigation of commonsense problem in large language models.
\newblock \emph{arXiv preprint arXiv:2303.16421}.

\bibitem[{Cao et~al.(2020)Cao, Dong, Wu, and Cheung}]{cao-etal-2020-factual}
Meng Cao, Yue Dong, Jiapeng Wu, and Jackie Chi~Kit Cheung. 2020.
\newblock \href {https://doi.org/10.18653/v1/2020.emnlp-main.506} {Factual error correction for abstractive summarization models}.
\newblock In \emph{Proceedings of the 2020 Conference on Empirical Methods in Natural Language Processing (EMNLP)}, pages 6251--6258, Online. Association for Computational Linguistics.

\bibitem[{Chang et~al.(2023)Chang, Wang, Wang, Wu, Zhu, Chen, Yang, Yi, Wang, Wang et~al.}]{chang2023survey}
Yupeng Chang, Xu~Wang, Jindong Wang, Yuan Wu, Kaijie Zhu, Hao Chen, Linyi Yang, Xiaoyuan Yi, Cunxiang Wang, Yidong Wang, et~al. 2023.
\newblock A survey on evaluation of large language models.
\newblock \emph{arXiv preprint arXiv:2307.03109}.

\bibitem[{Chen et~al.(2023)Chen, Zhao, Yu, McKeown, and He}]{chen-etal-2023-relation}
Yanda Chen, Chen Zhao, Zhou Yu, Kathleen McKeown, and He~He. 2023.
\newblock \href {https://doi.org/10.18653/v1/2023.findings-emnlp.12} {On the relation between sensitivity and accuracy in in-context learning}.
\newblock In \emph{Findings of the Association for Computational Linguistics: EMNLP 2023}, pages 155--167, Singapore. Association for Computational Linguistics.

\bibitem[{Dai et~al.(2022)Dai, Dong, Hao, Sui, Chang, and Wei}]{dai-etal-2022-knowledge}
Damai Dai, Li~Dong, Yaru Hao, Zhifang Sui, Baobao Chang, and Furu Wei. 2022.
\newblock \href {https://doi.org/10.18653/v1/2022.acl-long.581} {Knowledge neurons in pretrained transformers}.
\newblock In \emph{Proceedings of the 60th Annual Meeting of the Association for Computational Linguistics (Volume 1: Long Papers)}, pages 8493--8502, Dublin, Ireland. Association for Computational Linguistics.

\bibitem[{De~Cao et~al.(2021)De~Cao, Aziz, and Titov}]{de-cao-etal-2021-editing}
Nicola De~Cao, Wilker Aziz, and Ivan Titov. 2021.
\newblock \href {https://doi.org/10.18653/v1/2021.emnlp-main.522} {Editing factual knowledge in language models}.
\newblock In \emph{Proceedings of the 2021 Conference on Empirical Methods in Natural Language Processing}, pages 6491--6506, Online and Punta Cana, Dominican Republic. Association for Computational Linguistics.

\bibitem[{Dong et~al.(2022)Dong, Li, Dai, Zheng, Wu, Chang, Sun, Xu, and Sui}]{dong2022survey}
Qingxiu Dong, Lei Li, Damai Dai, Ce~Zheng, Zhiyong Wu, Baobao Chang, Xu~Sun, Jingjing Xu, and Zhifang Sui. 2022.
\newblock A survey for in-context learning.
\newblock \emph{arXiv preprint arXiv:2301.00234}.

\bibitem[{Feng et~al.(2023)Feng, Ma, Yu, Huang, Wang, Chen, Peng, Feng, Qin et~al.}]{feng2023trends}
Zhangyin Feng, Weitao Ma, Weijiang Yu, Lei Huang, Haotian Wang, Qianglong Chen, Weihua Peng, Xiaocheng Feng, Bing Qin, et~al. 2023.
\newblock Trends in integration of knowledge and large language models: A survey and taxonomy of methods, benchmarks, and applications.
\newblock \emph{arXiv preprint arXiv:2311.05876}.

\bibitem[{Gao et~al.(2023)Gao, Dai, Pasupat, Chen, Chaganty, Fan, Zhao, Lao, Lee, Juan, and Guu}]{gao-etal-2023-rarr}
Luyu Gao, Zhuyun Dai, Panupong Pasupat, Anthony Chen, Arun~Tejasvi Chaganty, Yicheng Fan, Vincent Zhao, Ni~Lao, Hongrae Lee, Da-Cheng Juan, and Kelvin Guu. 2023.
\newblock \href {https://doi.org/10.18653/v1/2023.acl-long.910} {{RARR}: Researching and revising what language models say, using language models}.
\newblock In \emph{Proceedings of the 61st Annual Meeting of the Association for Computational Linguistics (Volume 1: Long Papers)}, pages 16477--16508, Toronto, Canada. Association for Computational Linguistics.

\bibitem[{Hu et~al.(2021)Hu, Shen, Wallis, Allen-Zhu, Li, Wang, Wang, and Chen}]{hu2021lora}
Edward~J. Hu, Yelong Shen, Phillip Wallis, Zeyuan Allen-Zhu, Yuanzhi Li, Shean Wang, Lu~Wang, and Weizhu Chen. 2021.
\newblock \href {http://arxiv.org/abs/2106.09685} {Lora: Low-rank adaptation of large language models}.

\bibitem[{Huang et~al.(2023)Huang, Shen, Zhang, Zhou, Rong, and Xiong}]{huang2023transformer}
Zeyu Huang, Yikang Shen, Xiaofeng Zhang, Jie Zhou, Wenge Rong, and Zhang Xiong. 2023.
\newblock Transformer-patcher: One mistake worth one neuron.
\newblock \emph{arXiv preprint arXiv:2301.09785}.

\bibitem[{Levy et~al.(2017)Levy, Seo, Choi, and Zettlemoyer}]{levy2017zero}
Omer Levy, Minjoon Seo, Eunsol Choi, and Luke Zettlemoyer. 2017.
\newblock Zero-shot relation extraction via reading comprehension.
\newblock \emph{arXiv preprint arXiv:1706.04115}.

\bibitem[{Li et~al.(2023)Li, Li, Song, Yang, Ma, and Yu}]{li2023pmet}
Xiaopeng Li, Shasha Li, Shezheng Song, Jing Yang, Jun Ma, and Jie Yu. 2023.
\newblock Pmet: Precise model editing in a transformer.
\newblock \emph{arXiv preprint arXiv:2308.08742}.

\bibitem[{Lin(2004)}]{lin2004rouge}
Chin-Yew Lin. 2004.
\newblock Rouge: A package for automatic evaluation of summaries.
\newblock In \emph{Text summarization branches out}, pages 74--81.

\bibitem[{Liu et~al.(2023{\natexlab{a}})Liu, Wang, Wang, Smith, Choi, and Hajishirzi}]{liu2023vera}
Jiacheng Liu, Wenya Wang, Dianzhuo Wang, Noah~A Smith, Yejin Choi, and Hannaneh Hajishirzi. 2023{\natexlab{a}}.
\newblock Vera: A general-purpose plausibility estimation model for commonsense statements.
\newblock \emph{arXiv preprint arXiv:2305.03695}.

\bibitem[{Liu et~al.(2023{\natexlab{b}})Liu, Zeng, He, Jiang, and He}]{liu2023makes}
Wei Liu, Weihao Zeng, Keqing He, Yong Jiang, and Junxian He. 2023{\natexlab{b}}.
\newblock What makes good data for alignment? a comprehensive study of automatic data selection in instruction tuning.
\newblock \emph{arXiv preprint arXiv:2312.15685}.

\bibitem[{Lu et~al.(2023)Lu, Yuan, Yuan, Lin, Lin, Tan, Zhou, and Zhou}]{lu2023instag}
Keming Lu, Hongyi Yuan, Zheng Yuan, Runji Lin, Junyang Lin, Chuanqi Tan, Chang Zhou, and Jingren Zhou. 2023.
\newblock \# instag: Instruction tagging for analyzing supervised fine-tuning of large language models.
\newblock \emph{arXiv e-prints}, pages arXiv--2308.

\bibitem[{Meng et~al.(2022{\natexlab{a}})Meng, Bau, Andonian, and Belinkov}]{meng2022locating}
Kevin Meng, David Bau, Alex Andonian, and Yonatan Belinkov. 2022{\natexlab{a}}.
\newblock Locating and editing factual associations in gpt.
\newblock \emph{Advances in Neural Information Processing Systems}, 35:17359--17372.

\bibitem[{Meng et~al.(2022{\natexlab{b}})Meng, Sharma, Andonian, Belinkov, and Bau}]{meng2022mass}
Kevin Meng, Arnab~Sen Sharma, Alex Andonian, Yonatan Belinkov, and David Bau. 2022{\natexlab{b}}.
\newblock Mass-editing memory in a transformer.
\newblock \emph{arXiv preprint arXiv:2210.07229}.

\bibitem[{Mitchell et~al.(2021)Mitchell, Lin, Bosselut, Finn, and Manning}]{mitchell2021fast}
Eric Mitchell, Charles Lin, Antoine Bosselut, Chelsea Finn, and Christopher~D Manning. 2021.
\newblock Fast model editing at scale.
\newblock \emph{arXiv preprint arXiv:2110.11309}.

\bibitem[{Mitchell et~al.(2022)Mitchell, Lin, Bosselut, Manning, and Finn}]{mitchell2022memory}
Eric Mitchell, Charles Lin, Antoine Bosselut, Christopher~D Manning, and Chelsea Finn. 2022.
\newblock Memory-based model editing at scale.
\newblock In \emph{International Conference on Machine Learning}, pages 15817--15831. PMLR.

\bibitem[{Reimers and Gurevych(2019)}]{reimers-gurevych-2019-sentence}
Nils Reimers and Iryna Gurevych. 2019.
\newblock \href {https://doi.org/10.18653/v1/D19-1410} {Sentence-{BERT}: Sentence embeddings using {S}iamese {BERT}-networks}.
\newblock In \emph{Proceedings of the 2019 Conference on Empirical Methods in Natural Language Processing and the 9th International Joint Conference on Natural Language Processing (EMNLP-IJCNLP)}, pages 3982--3992, Hong Kong, China. Association for Computational Linguistics.

\bibitem[{Sinitsin et~al.(2020)Sinitsin, Plokhotnyuk, Pyrkin, Popov, and Babenko}]{sinitsin2020editable}
Anton Sinitsin, Vsevolod Plokhotnyuk, Dmitriy Pyrkin, Sergei Popov, and Artem Babenko. 2020.
\newblock Editable neural networks.
\newblock \emph{arXiv preprint arXiv:2004.00345}.

\bibitem[{Thorne and Vlachos(2021)}]{thorne-vlachos-2021-evidence}
James Thorne and Andreas Vlachos. 2021.
\newblock \href {https://doi.org/10.18653/v1/2021.acl-long.256} {Evidence-based factual error correction}.
\newblock In \emph{Proceedings of the 59th Annual Meeting of the Association for Computational Linguistics and the 11th International Joint Conference on Natural Language Processing (Volume 1: Long Papers)}, pages 3298--3309, Online. Association for Computational Linguistics.

\bibitem[{Touvron et~al.(2023)Touvron, Martin, Stone, Albert, Almahairi, Babaei, Bashlykov, Batra, Bhargava, Bhosale et~al.}]{touvron2023LLaMA}
Hugo Touvron, Louis Martin, Kevin Stone, Peter Albert, Amjad Almahairi, Yasmine Babaei, Nikolay Bashlykov, Soumya Batra, Prajjwal Bhargava, Shruti Bhosale, et~al. 2023.
\newblock Llama 2: Open foundation and fine-tuned chat models.
\newblock \emph{arXiv preprint arXiv:2307.09288}.

\bibitem[{Wang et~al.(2023{\natexlab{a}})Wang, Cheng, Xu, Ding, Wang, and Zhang}]{wang2023evaluating}
Cunxiang Wang, Sirui Cheng, Zhikun Xu, Bowen Ding, Yidong Wang, and Yue Zhang. 2023{\natexlab{a}}.
\newblock Evaluating open question answering evaluation.
\newblock \emph{arXiv preprint arXiv:2305.12421}.

\bibitem[{Wang et~al.(2023{\natexlab{b}})Wang, Zhu, Liu, Zheng, Chen et~al.}]{wang2023knowledge}
Song Wang, Yaochen Zhu, Haochen Liu, Zaiyi Zheng, Chen Chen, et~al. 2023{\natexlab{b}}.
\newblock Knowledge editing for large language models: A survey.
\newblock \emph{arXiv preprint arXiv:2310.16218}.

\bibitem[{Wu et~al.(2023)Wu, Li, Xu, Dong, Wu, Bian, and Xiong}]{wu-etal-2023-depn}
Xinwei Wu, Junzhuo Li, Minghui Xu, Weilong Dong, Shuangzhi Wu, Chao Bian, and Deyi Xiong. 2023.
\newblock \href {https://doi.org/10.18653/v1/2023.emnlp-main.174} {{DEPN}: Detecting and editing privacy neurons in pretrained language models}.
\newblock In \emph{Proceedings of the 2023 Conference on Empirical Methods in Natural Language Processing}, pages 2875--2886, Singapore. Association for Computational Linguistics.

\bibitem[{Yao et~al.(2023)Yao, Wang, Tian, Cheng, Li, Deng, Chen, and Zhang}]{yao-etal-2023-editing}
Yunzhi Yao, Peng Wang, Bozhong Tian, Siyuan Cheng, Zhoubo Li, Shumin Deng, Huajun Chen, and Ningyu Zhang. 2023.
\newblock \href {https://doi.org/10.18653/v1/2023.emnlp-main.632} {Editing large language models: Problems, methods, and opportunities}.
\newblock In \emph{Proceedings of the 2023 Conference on Empirical Methods in Natural Language Processing}, pages 10222--10240, Singapore. Association for Computational Linguistics.

\bibitem[{Zhao et~al.(2023)Zhao, Zhou, Li, Tang, Wang, Hou, Min, Zhang, Zhang, Dong et~al.}]{zhao2023survey}
Wayne~Xin Zhao, Kun Zhou, Junyi Li, Tianyi Tang, Xiaolei Wang, Yupeng Hou, Yingqian Min, Beichen Zhang, Junjie Zhang, Zican Dong, et~al. 2023.
\newblock A survey of large language models.
\newblock \emph{arXiv preprint arXiv:2303.18223}.

\bibitem[{Zheng et~al.(2023)Zheng, Li, Dong, Fan, Wu, Xu, and Chang}]{zheng-etal-2023-edit}
Ce~Zheng, Lei Li, Qingxiu Dong, Yuxuan Fan, Zhiyong Wu, Jingjing Xu, and Baobao Chang. 2023.
\newblock \href {https://doi.org/10.18653/v1/2023.emnlp-main.296} {Can we edit factual knowledge by in-context learning?}
\newblock In \emph{Proceedings of the 2023 Conference on Empirical Methods in Natural Language Processing}, pages 4862--4876, Singapore. Association for Computational Linguistics.

\bibitem[{Zhou et~al.(2023)Zhou, Liu, Xu, Iyer, Sun, Mao, Ma, Efrat, Yu, Yu et~al.}]{zhou2023lima}
Chunting Zhou, Pengfei Liu, Puxin Xu, Srini Iyer, Jiao Sun, Yuning Mao, Xuezhe Ma, Avia Efrat, Ping Yu, Lili Yu, et~al. 2023.
\newblock Lima: Less is more for alignment.
\newblock \emph{arXiv preprint arXiv:2305.11206}.

\bibitem[{Zhu et~al.(2020)Zhu, Rawat, Zaheer, Bhojanapalli, Li, Yu, and Kumar}]{zhu2020modifying}
Chen Zhu, Ankit~Singh Rawat, Manzil Zaheer, Srinadh Bhojanapalli, Daliang Li, Felix Yu, and Sanjiv Kumar. 2020.
\newblock Modifying memories in transformer models.
\newblock \emph{arXiv preprint arXiv:2012.00363}.

\end{thebibliography}

\appendix

\section{Details of Evaluation}
\subsection{Details of Existing Metrics}
\label{app:metric}
There are three metrics based on logits mainly used to evaluate the performance of knowledge editing in previous work, namely Efficacy, Generalization, and Specificity.
\begin{itemize}
    \item \textbf{Efficacy} measures the accuracy of knowledge editing using \textbf{ES} (Efficacy Score) and \textbf{EM} (Efficacy Magnitude). For Simple type queries, the meaning of ES is $E\left[ I\left[ P(o^*) > P(o) \right] \right]$ , and EM is obtained by $E[P(o^*) - P(o)]$ .
    \item \textbf{Generalization} measures the accuracy of knowledge editing on Rephrase queries by using \textbf{RS} (Rephrase Score) and \textbf{RM} (Rephrase Magnitude). For Rephrase type queries, RS and RM are actually calculated to derive ES and EM under the condition of rephrasing queries.
    \item \textbf{Specificity} uses \textbf{NS} (Neighborhood Score) and \textbf{NM} (Neighborhood Magnitude) to measure the ability of knowledge editing to preserve unrelated knowledge. When dealing with OOS queries beyond the editing scope, no editing should take place, and the original facts should be preserved. Therefore, NS is obtained by $E\left[ I\left[ P(o) > P(o^*) \right] \right]$, and NM is obtained by $E[P(o) - P(o^*)]$ .
\end{itemize}

\subsection{Consistency with Human Evaluation}
\label{subapp:consistency}
\begin{table}[t]
    \centering
    \setlength\tabcolsep{4pt}
     \resizebox{0.48\textwidth}{!}{%
    \begin{tabular}{c|c|c}
    \hline
        Human Score & Auto Metric & Pearson Correlation \\ \hline
        \multirow{3}{*}{Editing} & TE & 0.7644 \\ 
         & SE & 0.7784 \\ 
        & Editing & 0.8074 \\ \hline
        \multirow{3}{*}{Retention} & TR & 0.9195 \\ 
        & SR & 0.8868 \\ 
         & Retention & 0.9255 \\ \hline
        \multirow{3}{*}{Overall} & Editing & 0.5356 \\ 
         & Retention & 0.7612 \\ 
         & Overall & 0.839 \\ \hline
\end{tabular}
}
\caption{The Pearson correlation coefficient between auto metrics and manual scores. For the auto metrics, Editing is the average of TE and SE; Retention is the average of TR and SR; Overall is the average of Editing and Retention.}
\label{tab:human_eval}
\end{table}
In Section \ref{subsec:eval_framework}, we proposed a comprehensive evaluation framework, incorporating editing metrics (TE, SE) and retention metrics (TR, SR) to evaluate the quality of output text after knowledge editing. Prior to employing these metrics for evaluation, it was imperative to ensure their validity and necessity. To address this, we sample 300 data points from the test set (comprising Simple, Rephrase, and OOS examples in a 1:1:1 ratio) and enlist human evaluators to independently score them from the perspectives of editing,  retention, and overall assessment. 

The rules for human scorers scoring the effectiveness of knowledge editing are as follows: in terms of editing, for INS queries, scoring is as follows: 0 points if there is no editing at all; 0.5 points if there are partial edits, and the sentence still retains old knowledge or exhibits logical inconsistencies; 1 point for perfect knowledge editing with no issues. For OOS queries, the scoring rules are reversed. In the retention aspect, after disregarding content related to the edited knowledge in the sentence, for responses within the editing scope: 0 points for very poor consistency between new and old responses; 0.5 points for ordinary consistency; 1 point for excellent consistency. In the overall aspect, human scorers are required to consider the overall impact of knowledge editing and assign scores within the range of 0, 1, 2, 3, 4 to the edited outputs. Then, we conduct Pearson correlation analyses between these human scores and our automated metrics.

As shown in Table \ref{tab:human_eval}, both textual metrics (TE, TR) and semantic metrics (SE, SR) demonstrate commendable consistency scores with human ratings, affirming the effectiveness of the proposed metrics. Moreover, Whether for editing or retention, the consistency score of the joint assessment of textual and semantic dimensions surpasses that of any individual metric. This underscores the necessity of incorporating both textual and semantic metrics in the evaluation process.
Finally, the Pearson correlation coefficient between auto editing and human overall score is a mere 0.5356. However, a combined evaluation of editing and retention metrics yield a significantly higher consistency score of 0.839 with human judgments. This suggests that effective alignment with human preferences cannot rely solely on editing scores but requires a comprehensive assessment integrating both editing and retention metrics.

\subsection{Pseudo-code of Evaluation Framework}
\label{subapp:evalutaion_detail}
We summarize the pseudo-code of our proposed evaluation framework in Algorithm \ref{alg:evaluation}.

\section{Details of Method}
\subsection{Pseudo-code of PostEdit}
\label{subapp:postedit}
We summarize the pseudo-code for training post-editor and inference of postEdit in Algorithm \ref{alg:post-editor} and Algorithm \ref{alg:postEdit}, respectively.

\subsection{Details of Prompts}
\label{subapp:prompt}
We demonstrate the two prompt templates $T^{aug}$ and $T^{edit}$ used in the postEdit method as follows:

\begin{tcolorbox}[
colback=white!10!white,
colframe=black!75!black,
title=Prompt Template $T^{aug}$,
breakable]
For the following query and original response, you need to follow in order:\\Firstly, locate all spans related to the \textbf{old fact:\{$\textbf{s}$\} \{$\textbf{r}$\} \{$\textbf{o}$\}} in original reply;\\Secondly, modify these spans according to \textbf{new fact: \{$\textbf{s}$\} \{$\textbf{r}$\} \{$\textbf{o}^*$\}}.\\Thirdly, output the edited response based on the modified spans (Do not output other content).\\\#\#\# The query:\\\{$x$\}\\\#\#\# Original response:\\\{$y_o$\}\\\#\#\# Edited response:
\end{tcolorbox}

\begin{tcolorbox}[
colback=white!10!white,
colframe=black!75!black,
title=Prompt Template $T^{edit}$,
breakable]
\label{response-aug prompt}
\#\#\# Instruction:\\You will assume the role of an editor. For the following query and original response, if the new fact impacts the query or original response, incorporate the new fact into the original response. If not, simply output the following word: retain.\\\#\#\# New fact:\\The answer of \{$s$\} \{$r$\} has been updated from \{$o$\} to \{$o^*$\}.\\\#\#\# The query:\\\{$x$\}\\\#\#\# Original response:\\\{$y_o$\}\\\#\#\# Edited response:
\end{tcolorbox}


\begin{table*}[t]
    \centering
 \resizebox{0.9\textwidth}{!}{%
    \begin{tabular}{c|cccc}
         \hline \textbf{Dataset}& \textbf{Data Type} & \textbf{Train Number} & \textbf{Test Number} & \textbf{Length of Original Response (mean/max)} \\
         \hline \multirow{4}{*}{CounterFact}&  ALL& 30000 & 1500 & 51.34/436\\
         & Simple & 10000 &  500& 50.40/436\\
         &Rephrase  & 10000 &  500& 53.03/374\\
         & OOS &10000  &  500& 50.59/367\\
         \hline \multirow{4}{*}{zsRE}  &ALL  &  30000&1500&22.39/406 \\
         & Simple &  10000& 500 & 14.84/119\\
         & Rephrase &10000  & 500 &18.38/257 \\
         & OOS & 10000 &500  & 33.96/406\\
         \hline
    \end{tabular}
}
    \caption{Statistical information on the sampled datasets.}
    \label{tab:dataset}
\end{table*}

\begin{algorithm*}[t]
\caption{Pseudo-code of Evaluation Framework in a Python-like style.}
\label{alg:evaluation}
\SetKwInput{KwInput}{Input}                
\SetKwInput{KwOutput}{Output}              
\DontPrintSemicolon
\SetKwComment{Comment}{\#}
\Comment{\textcolor[RGB]{63, 127, 127}{\\\# x: the input of LLM (All text is processed in lowercase, the same below.)\\\# x\_label: "INS" if x in editing scope else "OOS"\\\# y\_o, y\_e: the original 
 and edited output of LLM\\\# o\_old, o\_new: the object of old knowledge $t$ and new knowledge $t^*$ for editing\\\# k\_old, k\_new: text format of $t$ and $t^*$\\\# k\_self: text format of LLM's self-knowledge $t_o$ and is equivalent to [x, y\_o]\\\# func\_entail(a,b): return True if a entails b else False by using a NLI model\\\# func\_rouge(a,b): return the ROUGE socre of a and b\\\# func\_sim(a,b): return the similarity of a and b using a SBERT model\\\ }}

  \SetKwFunction{FMain}{Main}
  \SetKwFunction{FTE}{\textbf{TE}}
  \SetKwFunction{FSE}{\textbf{SE}}
  \SetKwFunction{FTR}{\textbf{TR}}
  \SetKwFunction{FSR}{\textbf{SR}}
 
  \SetKwProg{Fn}{def}{:}{}
  \Fn{\FTE{\textnormal{y\_e, x\_label, o\_old, o\_new}}}{
        ctn\_old=1 \textbf{if} o\_old \textbf{in} y\_e \textbf{else} 0\;
        ctn\_new=1 \textbf{if} o\_new \textbf{in} y\_e \textbf{else} 0\;
        \textbf{if} x\_label=="INS":\;
            \qquad TE\_score=0.5*ctn\_new + 0.5*(1-ctn\_old)\;
        \textbf{else}:\;
            \qquad TE\_score=0.5*ctn\_old + 0.5*(1-ctn\_new)\;
        \textbf{return} TE\_score
  }
  \;

  \SetKwProg{Fn}{def}{:}{}
  \Fn{\FSE{\textnormal{x\_label, x, y\_e, k\_old, k\_new, k\_self, func\_entail}}}{
    ent\_new=1 \textbf{if} func\_entail(x+" "+y\_e,k\_new) \textbf{else} 0\;
    \textbf{if} x\_label=="INS":\;
        \qquad ent\_old=1 \textbf{if} func\_entail(x+" "+y\_e,k\_old) \textbf{else} 0\;
        \qquad SE\_score=0.5 * ent\_new + 0.5 * (1-ent\_old)\;
    \textbf{else}:\;
        \qquad ent\_old=1 \textbf{if} func\_entail(x+" "+y\_e,k\_self) \textbf{else} 0\;
        \qquad SE\_score=0.5*ent\_old + 0.5*(1-ent\_new)\;
    \textbf{return} SE\_score
  }
  \;

  \SetKwProg{Fn}{def}{:}{}
  \Fn{\FTR{\textnormal{x\_label, y\_o, y\_e, o\_old, o\_new, func\_rouge}}}{
    \textbf{if} x\_label=="INS":\;
        \qquad TR\_score=func\_rouge(y\_o.replace(o\_old,"mask"),
                            y\_e.replace(o\_new,"mask"))\;
    \textbf{else}:\;
        \qquad TR\_score=func\_rouge(y\_o,y\_e)\;
    \textbf{return} TR\_score
  }
  \;
  \SetKwProg{Fn}{def}{:}{}
  \Fn{\FSR{\textnormal{x\_label, y\_o, y\_e, o\_old, o\_new, func\_sim}}}{
    \textbf{if} x\_label=="INS":\;
        \qquad SR\_score=func\_sim(y\_o.replace(o\_old,"mask"),
                        y\_e.replace(o\_new,"mask"))\;
    \textbf{else}:\;
        \qquad SR\_score=func\_sim(y\_o,y\_e)\;
    \textbf{return} SR\_score
  }
\end{algorithm*}

\section{Details of Experiments}
\subsection{Details of Datasets}

\label{subapp:dataset}

In this work, we mainly used two datasets: zsRE and CounterFact.
\begin{itemize}
\item \textbf{zsRE} \cite{levy2017zero} is one of the most popular question answering  
 (QA) datasets which use question rephrasing as the equivalence neighborhood. These queries of Rephrase type are generated by back-translation. In zsRE, the relationship between entities is associated with a set of crowd-sourced generated questions. Additionally, zsRE associates questions with randomly generated sentences to add out-of-editing scope examples.
\item \textbf{CounterFact} \cite{meng2022locating} is a more challenging dataset than zsRE, the expected output of which is contradictory to the fact. It is built to distinguish superficial alterations in the word selections and significant, generalized modifications in its foundational factual knowledge. In CounterFact, the edited answer to the question can sometimes be counterfactual to real world, which makes it harder for the model to predict desired answer and avoid the effects of pre-trained LLMs knowing these desired facts before editing.
\end{itemize}
Following the previous work \cite{zheng-etal-2023-edit}, for CounterFact, we designate data with edit id numbers ranging from 0 to 2000 as the test set for knowledge edit, while the remaining data constitute the training set. As we adopt ChatGPT as our base LLM in main experiments, in order to control the dataset size, we randomly sampled 30,000 examples (10,000 each for Simple, Rephrase, and OOS) from the original training set. These samples constitute our training set. Additionally, we randomly selected 1,500 examples (500 each for Simple, Rephrase, and OOS) from the original test set to create our query test set. The original response for INS test queries are ensured to hit the old knowledge object before editing, and the OOS are ensured to have no wrong knowledge before editing. We present the statistical information of the datasets after sampling in Table \ref{tab:dataset},  and show a training sample and test sample from zsRE respectively as follows:

\begin{tcolorbox}[
colback=white!10!white,
colframe=black!75!black,
title=Sample From zsRE Training Set,
breakable]
\{\

        \hspace{0.5cm}\textbf{"edit\_id":} 15000,\
        
        \hspace{0.5cm}\textbf{"edit":} "Denis Dyack  >>  Denys de La Tour || Who is the designer of Too Human?",\
        
        \hspace{0.5cm}\textbf{"query":} "Who is the designer from Too Human?",\
        
        \hspace{0.5cm}\textbf{"query\_type":} "rephrase",\
        
        \hspace{0.5cm}\textbf{"original\_response\_by\_gpt3.5":} "The designer of Too Human is Denis Dyack.",\
        
        \hspace{0.5cm}\textbf{"edited\_response\_by\_gpt4":} "The designer of Too Human is Denys de La Tour."\
\}\        
\end{tcolorbox}

\begin{tcolorbox}[
colback=white!10!white,
colframe=black!75!black,
title=Sample From zsRE Test Set,
breakable]
\{\

        \hspace{0.5cm}\textbf{"edit\_id":} 70,\
        
        \hspace{0.5cm}\textbf{"edit":} "Serpens  >>  Andromeda || Which constellation is NGC 6604 in?",\
        
        \hspace{0.5cm}\textbf{"query":} "Which constellation does NGC 6604 belong to?",\
        
        \hspace{0.5cm}\textbf{"query\_type":} "rephrase",\
        
        \hspace{0.5cm}\textbf{"original\_response":} "NGC 6604 belongs to the constellation of Serpens."\
        
\}\        
\end{tcolorbox}

\subsection{Details of Baselines}
\label{subapp:baselines}
\begin{itemize}
    \item \textbf{IKE} \cite{zheng-etal-2023-edit} is a method of knowledge editing that does not involve modifying the parameters of LLMs. It defines three types of demonstration formatting templates including copy, update, and retain. These templates serve distinct functions and act as guiding principles for the language model, enabling it to edit knowledge through in-context learning, allowing IKE to maintain both efficiency and excellent generalization and specificity. This opens up the possibility of employing IKE for the task of knowledge editing even in scenarios involving black-box models.
    \item \textbf{PROMPT} \cite{zheng-etal-2023-edit} is similar to IKE, as a method of knowledge editing through in-context learning. However, unlike IKE, PROMPT doesn't require constructing three types of demonstrations but directly provides new knowledge to the LLM for knowledge editing.
    \item \textbf{SERAC} \cite{mitchell2022memory} is a memory-based method of knowledge editing. This method stores edits in explicit memory and learns to reason about these edits as needed to adjust the predictions of the base LLM without modifying parameters. SERAC uses an explicit cache of user-provided edit descriptors, alongside a scope classifier and surrogate model. When presented with a query, SERAC uses the scope classifier to determine if the query falls within the editing scope. If it does, the output is predicted via the surrogate model; otherwise, it defers to the base LLM for the output.
    \item \textbf{SERAC (ChatGPT)} In SERAC, the surrogate model is obtained by fine-tuning a smaller language model compared to the base LLM. We utilize ChatGPT as the surrogate model to derive a SERAC variant that requires no additional training.
\end{itemize}

\subsection{Details of Implementation}
\label{subapp:Implementation}
As described in Section \ref{subsec:eval_framework}, our evaluation framework employs a NLI model for computing SE, ROUGE scores for computing TR, and a SBERT model for computing SR. In details, SE utilizes albert-xxlarge-v2-snli\_mnli\_fever\_anli\_R1\_R2\_R3-nli\footnote{https://huggingface.co/ynie/albert-xxlarge-v2-snli\_mnli\_fever\_anli\_R1\_R2\_R3-nli} as the NLI model; ROUGE score is implemented through the rouge library\footnote{https://pypi.org/project/rouge}, using the F1 score of ROUGE-1; SR uses all-MiniLM-L6-v2\footnote{https://huggingface.co/sentence-transformers/all-MiniLM-L6-v2} as the SBERT model.

For training of post-editor, we employ ChatGPT (gpt-3.5-turbo-0301) for original response augment and GPT-4 (gpt-4-0613) for edited response augment \footnote{https://platform.openai.com/docs/models}, with the default temperature coefficient ($t=0.1$). In order to enhance training efficiency and reduce the number of updated parameters, we adopt the LoRA strategy \cite{hu2021lora} to finetune LLaMA 2-7B.  Specifically, the rank of LoRA is set to 8, with $lora\_alpha$ at 16 and $lora\_dropout$ at 0.05. The LoRA update matrix is applied to the self-attention and FFN layers, with $target\_modules$ as ["q\_proj","k\_proj","v\_proj","o\_proj","gate\_proj",\\"down\_proj","up\_proj"]. 
We train 5 epochs to optimize post-editor, employing a batch size of 128 and a learning rate of 5e-2. We also use the warmup and cosine annealing strategy, with a warmup ratio of 0.1 and the  Adam optimizer. 

For retriever of postEdit, consistent with all baselines, we use all-MiniLM-L6-v2 to encode queries and edit knowledge, while employing dot product as the similarity function. For base LLM, we use ChatGPT (gpt-3.5-turbo-0301)  in main experiments, with a temperature coefficient of 0.1. During inference of post-editor, we set the  temperature coefficient of 0.1 and use beam search to decode the output,  where $num\_beams$ is set to 4. To further improve the inference speed, we apply 8-bit quantization when loading post-editor.

In terms of baselines, for SERAC, we fine-tune the surrogate model using the same LLAMA2-7B as post-editor and the similarity discrimination threshold is set at 0.7, determined through hyperparameter search on the training set (ranging from 0.1 to 0.9 with a step size of 0.1). To better maintain consistency between baselines and postEdit implementations, we adopt training output targets consistent with postEdit for the surrogate model of SERAC, i.e., GPT-4 augmented edited response, rather than new objects of editing knowledge, aiming to achieve higher stylistic retention. For IKE, we set the number of demonstration examples to 32. The rest of the hyperparameter settings for the baselines follow the default configurations in their original papers. All experiments use  a single Nvidia A100 GPU (80 GB of memory).

\SetKw{KwRequire}{Require: }
\SetKwComment{Comment}{$\triangleright$}{}
\DontPrintSemicolon
\begin{algorithm*}[t]
\SetAlgoLined
\KwData{training dataset $D_{train}=\{(e_i,x_i)\}$}
\KwRequire{\textnormal{ base LLM $f_{base}$, GPT-4 $f_{gpt4}$, trainable generative model $f_{edit}$, training epoch $\textbf{E}$, batch size $\textbf{B}$}}\;
\For{$i$ in $1,\cdots,|D_{train}|$}{
$y^{aug}_{i,o}=f_{base}(x_i)$\Comment*[r]{\textbf{\textrm{Original Response Augment}}}
\eIf{$x_i \in \mathcal{X}_e$}
{$y^{aug}_{i,e}=f_{gpt4}(T^{aug}(e_i,x_i,y^{aug}_{i,o}))$\Comment*[r]{\textbf{\textrm{Edited Response Augment}}}
\If{$\operatorname{TE}(y^{aug}_{i,e})\neq1$ or $\operatorname{SE}(y^{aug}_{i,e})\neq1$}{\textbf{delete} $(e_i,x_i,y^{aug}_{i,o},y^{aug}_{i,e})$\;}}
{$y^{aug}_{i,e}=\left<Retain\right>$\;}
}
$D^{aug}_{train}=\{(e_i,x_i,y^{aug}_{i,o},y^{aug}_{i,e})\}$\;
\For{$epoch$ in $1,\cdots,\textbf{E}$}{
    \For{$iter$=$0, 1, 2,\cdots$}{
        sample a mini-batch $\textbf{B}$ from $D^{aug}_{train}$\Comment*[r]{\textbf{\textrm{Supervised Fine-tuning}}}
        compute $\mathcal{L}_{sft}$ by equation \ref{eq:sft} and optimize $f_{edit}$
}
}
\KwOut{trained post-editor $f_{edit}$}
 \caption{Train post-editor}
 \label{alg:post-editor}
\end{algorithm*}

\SetKw{KwRequire}{Require: }
\DontPrintSemicolon
\begin{algorithm*}[t]
\SetAlgoLined
\KwIn{use query $x$}
\KwRequire{\textnormal{Editing Memory $M_e$, base LLM $f_{base}$, post-editor $f_{edit}$, SBERT retriever $f_{retr}$}}\;
get original response: $y_o=f_{base}(x)$\;
retrieve the most similar edit index: $i^*=\operatorname{argmax}_{0\leq i<|M_e|}\ \operatorname{sim}(x,e_i)$\;
get post-editor's output: $f_{edit}(x_{edit})=f_{edit}(T^{edit}(e_{i^*},x,y_o))$\;
\eIf{$f_{edit}(x_{edit})\neq\left<Retain\right>$}{
$y_e=f_{edit}(x_{edit})$\;
}{
$y_e=y_o$\;
}
\KwOut{final response $y_e$}
 \caption{Inference of PostEdit}
 \label{alg:postEdit}
\end{algorithm*}

\end{document}